# Machine Learning for Reliability Engineering and Safety Applications: Review of Current Status and Future Opportunities

Zhaoyi Xu, Joseph Homer Saleh[*]

*School of Aerospace Engineering, Georgia Institute of Technology, USA*

Machine learning (ML) pervades an increasing number of academic disciplines and industries. Its impact is profound, and several fields have been fundamentally altered by it, autonomy and computer vision for example; reliability engineering and safety will undoubtedly follow suit. There is already a large but fragmented literature on ML for reliability and safety applications, and it can be overwhelming to navigate and integrate into a coherent whole. In this work, we facilitate this task by providing a synthesis of, and a roadmap to this ever-expanding analytical landscape and highlighting its major landmarks and pathways. We first provide an overview of the different ML categories and sub-categories or tasks, and we note several of the corresponding models and algorithms. We then look back and review the use of ML in reliability and safety applications. We examine several publications in each category/sub-category, and we include a short discussion on the use of Deep Learning to highlight its growing popularity and distinctive advantages. Finally, we look ahead and outline several promising future opportunities for leveraging ML in service of advancing reliability and safety considerations. Overall, we argue that ML is capable of providing novel insights and opportunities to solve important challenges in reliability and safety applications. It is also capable of teasing out more accurate insights from accident datasets than with traditional analysis tools, and this in turn can lead to better informed decision-making and more effective accident prevention.

---

[*] Corresponding author. Tel: + 1 404 385 6711; fax: + 1 404 894 2760
E-mail address: jsaleh@gatech.edu (J. H. Saleh)

# 1. Introduction

This work provides a synthesis of and a roadmap to the growing and diverse literature on Machine Learning (ML) for reliability engineering and safety applications. It also outlines future opportunities and challenges for ML in these areas of applications.

Machine learning pervades an increasing number of fields, from banking and finance [1, 2, 3] to healthcare [4, 5, 6], robotics [7, 8, 9], transportation [10, 11, 12], e-commerce [13, 14, 15], and social networks to mention a few. Few academic disciplines are likely to remain immune from its influence. Its impact is profound, and it will continue to upend traditional academic disciplines and industries. This quiet but relentless wave of creative disruption was enabled in part by the advent of *big data*, the collection and storage of the massive amount of data, and the development of powerful models and algorithms to probe it. Roughly speaking, "machine learning [is] a set of methods that can detect patterns in data, and then use the uncovered patterns to predict future data, or to perform other kinds of decision making under uncertainty" [16]. Machine learning is "essentially a form of applied statistics with increased emphasis on the use of computers to statistically estimate complicated functions and a decreased emphasis on proving confidence intervals around these functions" [17]. Several academic fields have been fundamentally altered by machine learning, controls and autonomy or computer vision for example, reliability engineering and safety analysis will undoubtedly follow suit. There is already a large but fragmented literature on ML for reliability and safety applications, and it can be overwhelming to navigate. We propose to facilitate this task in this work by describing this ever-expanding analytical landscape and highlighting its major landmarks and pathways.

In a previous review work of reliability engineering [18], the author summarized the progress to date in the field and highlighted some of its challenges. He noted, for example, that new reliability analysis tools are required to model the ever-increasing complexity of systems being designed. He also highlighted the need for better predictions and uncertainty propagation in reliability and safety analysis. In support of this observation, Li et al. [19] pointed out that traditional tools for lifetime predictions of engineering components are becoming less capable of meeting the industry's increasing demand for precision and accuracy. Although these topics will be discussed in the next sections, we simply note here that ML is well suited to address these challenges. ML is capable of providing new insights and opportunities to solve important challenges in reliability engineering and safety analysis. It is also capable of teasing out more accurate insights from accident datasets, or degradation and survival data for example that were beyond the capability of traditional analysis tools. This, in turn, can lead to better decision-making for the design and operation of engineering systems and more effective accident prevention.

Our first objective in this work is to provide a synthesis of, and a roadmap to current ML use in reliability engineering and safety applications. Our second objective is to outline some future opportunities for ML in reliability and safety applications. Overall, we hope this work encourages the reliability and safety communities to upgrade their traditional analytical toolkits to include (different aspects of) ML and leverage its significant potential.

The remainder of the article is organized as following. In section 2, we provide a brief overview of ML and its main categories and sub-categories or tasks. In section 3, we review the current status of ML use in reliability and safety, and we provide a synthesis of and a roadmap to this diverse literature. In section 4, we discuss future opportunities for ML in reliability and safety applications. Finally, we conclude this work in section 5.

## 2. Machine learning: a brief overview

Machine learning, as noted previously, is a set of methods for learning from data and uncovering patterns in it. This learning, in turn, can be put to use for inferential and prediction purposes, or to perform some decision-making under uncertainty. Given the type of data available and the kind of questions being asked for understanding it, different categories of ML are available for the task at hand. ML can be classified into three major categories, with a fourth one straddling the first two: (i) supervised learning; (ii) unsupervised learning; (iii) semi-supervised learning; and (iv) reinforcement learning. A tree diagram of this categorization along with the common sub-categories is shown in Fig. 1. A brief introduction to these categories is provided next.

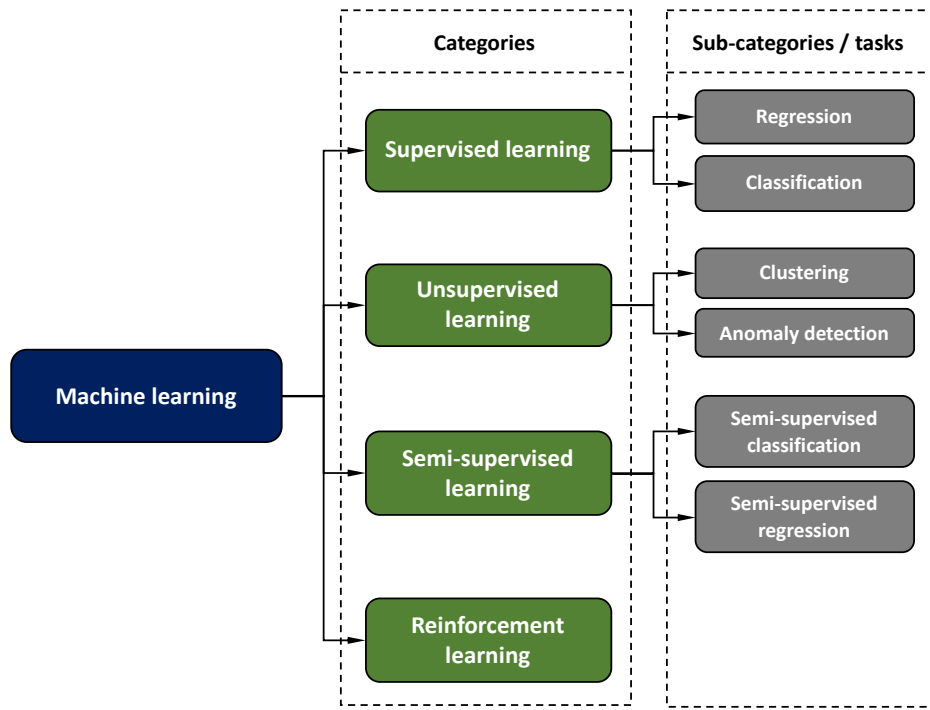

**Figure 1**. Machine learning categories and sub-categories or tasks, the latter are not meant to be exhaustive. Anomaly detection algorithms exist in supervised and semi-supervised learning modes; the more widely used are in unsupervised mode. Semi-supervised regression is a smaller area of research that its classification counterpart. Not shown here is Active learning for example, in which the learning algorithm can query the user to provide labels to carefully selected previously unlabeled data.

The process of generating and collecting data is essential to science and engineering, and it is becoming increasing more so to all other fields. The different categories of machine learning offer ways for understanding different aspects of what data is collected. Consider, for example, a dataset with thousands of observations about the main rotor diameters of helicopters, their Maximum Takeoff Weight (MTOW), number of engines (single or twin), and other features. Assume we want to determine, for instance, if there is a relationship between the main rotor diameter and the other helicopter features. And if so, how can we quantify it, and how accurately can we predict the former from the latter? In this case, we have paired observations with the output of interest Yi, here the main rotor diameter, and a vector of features Xi, here MTOW and number of engines (Xi, Yi). In machine learning, the output of interest Yi is referred to as the label, and it is also known as the response or dependent variable. The inputs available Xi are known as features, covariates, or predictors. One way of understanding the different categories of ML is in relation to Fig. 2. The unknown function $f(X)$ represents the information that the feature vector X provides about the label Y. Roughly speaking, machine learning is about estimating this function with $\hat{f}$.[1]

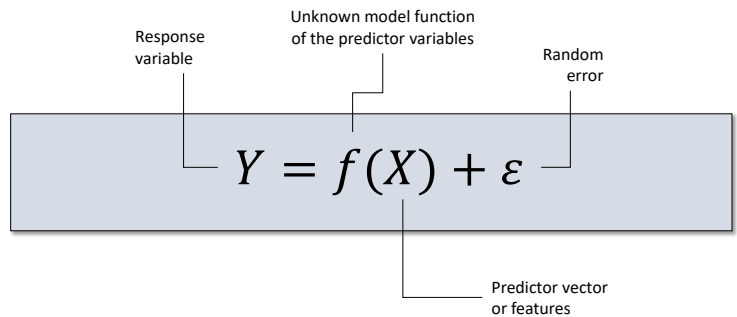

**Figure 2**: Ingredients of machine learning

What is "learned" from the data is the model $\hat{Y} = \hat{f}(X)$. In this expression, $\hat{Y}$ is the predicted response variable given the predictors $X$, along with in some cases the uncertainty quantification associated with the prediction. Supervised learning consists in estimating $\hat{f}$ when the dataset includes paired features and labels, that is Xi and Yi. Unsupervised learning consists in estimating $\hat{f}$ when the dataset is unlabeled, that is, there is no Yi. The dataset in this case consists of only input variables. This is akin to searching blindly for patterns in a dataset without a guide to supervise the learning, hence the qualifier unsupervised. For this reason, unsupervised learning is also known as knowledge discovery, and it is more challenging than supervised learning. Semi-supervised learning, as can be surmised, consists in estimating $\hat{f}$ when only partial labels are available, that is, some paired observations are available, but not all the features have labels associated with them (some Xi have no Yi). Reinforcement learning does not directly fit into this framework, but it roughly consists in exploring for $X$ and searching for the optimal $X^*$ given a desired output $Y$. More details about each category are discussed in the next subsections.

## 2.1 Supervised learning (SL)

Supervised learning, as noted previously, consists in estimating a function that maps an input vector to an output given a dataset of paired observations:

$$\hat{f}: X \rightarrow Y$$

Supervised learning is done for inferential or prediction purposes. In the inferential case, we seek to understand for example, the strength of the association between the predictors and the response variable, or how the latter varies with the former. For prediction purposes, we seek to predict the responses for new observed features not in the dataset. Given the nature of the response variable Y, a quantitative or qualitative variable, two major sub-categories of supervised learning are available: regression and classification.

---

[1] This can be parametric or non-parametric.

Regression problems consist in estimating $\hat{f}$ for which the response variable is quantitative, such as blood pressure, temperature, weight, or cost for example. To carry out this estimation, several machine learning models and algorithms are available, from the widely used basic linear regression (LR) [20] and polynomial response surface (PRS) [21], to more advanced techniques, such as support vector regression (SVR) [22], decision tree regression (DTR) [23], and random forest regression (RFR) [24]. Some ML models, such as Gaussian Process Regression (GPR) and Bayesian network, offer a distinctive advantage by providing accurate uncertainty quantification in regression problems [25, 26]. Furthermore, with the recent development of deep learning techniques, deep neural network (DNN) models are becoming increasingly popular with regression for high-dimensional, nonlinear problems. DNNs can be more accurate, and they are better at regulating the overfitting problem than their traditional ML counterparts [27, 28].

The second sub-category of supervised learning shown in Fig. 1, classification, involves a response variable Y that is a qualitative, for example 0/1 or eye colors. Classification problems consists again in estimating $\hat{f}$ given a labeled dataset, then using this learned knowledge, for example, to predict the classification probability of new observed features. The problem may involve an output with only two classes or bi-class, such as spam email identification (spam/not spam) [29], or multi-class output, such as speech and handwriting recognition [30]. Commonly used ML methods for classification problems include logistic regression [31], k-nearest neighbor (KNN) [32], support vector machine (SVM) [16], decision tree (DT) [16], boosted tree [33], and random forest (RF) [34]. An excellent reference for a wide range of classification models can be found in Ref. [35]. As with regression tasks, DNNs models are also used for classification, and with proper activation and loss functions, they provide better performance than their traditional ML counterparts [16, 36]. We will provide more details and examine the use of supervised learning in reliability engineering and safety application in section 3.

## 2.2 Unsupervised learning (USL)

Unsupervised learning consists in examining datasets with only input variables $\mathbf{X}_i$ and no corresponding label or response $Y_i$. This problem setup can be disconcerting at first, and it is fair to inquire about its objective and what can be "learned" in this situation. We first note that unlike supervised learning, unsupervised learning is not concerned with predicting an output variable since none is available. Its objective instead is to explore the feature space $\mathbf{X}_i$ and find patterns in the dataset. Two major sub-categories or tasks of unsupervised learning are available based on the nature of the patterns sought: clustering and anomaly detection.

Clustering, also known as unsupervised classification, consists in dividing the observations into clusters that share some similarities in the feature space. Clustering and (supervised) classification share on the surface some similarities in that they both deal with observations in groups; except in supervised classification the groups or classes are predefined, and the observations are assigned to different classes before the analysis starts (i.e., labeled data). Whereas in clustering, neither the groups nor their numbers are known beforehand, and the assignment of the unlabeled data to specific meaningful clusters is the main outcome of the analysis. Choosing the number of clusters, deciding what constitute similarity in the $\mathbf{X}_i$ space (similarity measures), and selecting the evaluation criteria to assess the quality of the output (e.g., intra-cluster homogeneity and inter-cluster separability) are typical challenges of clustering. Clustering methods are widely used in a host of applications, from image segmentation and object recognition, to document retrieval/data mining, genomics, and countless e-commerce applications. There is a significant number of clustering algorithms to choose from, and it can be overwhelming for the user to navigate the thicket of methods available for this task. Two excellent reviews of clustering methods can be found in Jain et al. [37] and Saxena et al. [38]. Clustering methods include the widely used K-means algorithm [39], a variety of hierarchical clustering methods [40], density-based clustering (DBC) [41], and Gaussian mixing model (GMM) [16]. With the recent development of DNN, there is a growing focus on combining deep learning feature extraction models such as deep auto-encoder, with clustering algorithms to handle problems with high-dimensional noisy data [42, 43].

The second sub-category of unsupervised learning shown in Fig. 1, anomaly detection, consists in identifying unexpected observations in a dataset. The term *anomaly* in this contest is used in a broader sense than how it is understood in the reliability and safety communities. For example, a person who steals your credit card will likely make purchases that derive from a different probability distribution of purchases than your own. Anomaly detection, as understood in machine learning, will flag these purchases as likely fraudulent. More generally, anomaly detection in ML refers to "the problem of finding patterns in data that do not conform to expected normal behavior" [44]. Anomaly detection algorithms have found a vast range of applications in many domains because they often produce critical information that can be acted upon and prompt meaningful intervention. For example, anomaly detection is used in cyber-security and intrusion detection [45], in banking and insurance fraud detection, in medical applications [46], and increasingly in reliability and safety applications as will be discussed in section 3. Two generically different types of anomalies are worth distinguishing, and they lead to different types of algorithms to detect them: (i) *point anomalies*, and (ii) *contextual anomalies*. The simplest are point anomalies, and they are defined as individual observations that can be assessed as anomalous with respect to other data points. The left panel in Fig. 3 illustrates point anomalies within a 2D feature space. Non-anomalous data occur in dense neighborhoods, whereas point anomalies occur in low density regions far from their closest neighbors or established clusters, as seen in Fig. 3. Contextual anomalies are more interesting and more difficult to handle than point anomalies. If a data is anomalous in a given context, but not so when evaluated in isolation, it is termed a contextual anomaly. The context can be broadly defined, and it is often specified in terms of spatial data or time-series data. For example, the right panel in Fig. 3 illustrates a contextual anomaly in a time-series.

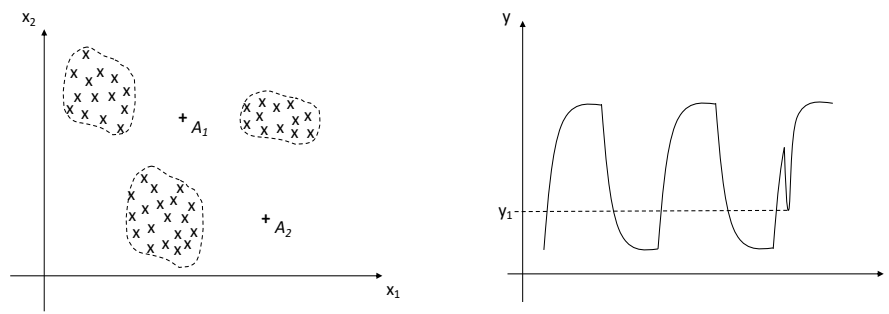

**Figure 3**. Illustrations of two types of anomalies: point anomalies (left panel; $A_1$ and $A_2$ are point anomalies), and contextual anomalies (right panel; $y_1$ is a contextual anomaly, its value is within the range of normal values except it is *anomalous in context*, here the context is derived from the time-series data). Adapted from Ref. [44].

Data for anomaly detection can come in streaming fashion, in which case it requires online analysis, or the entire dataset can be available at the onset, and the analysis carried out offline. These lead to different types of algorithms for which computational efficiency is paramount (in the streaming case) or not. Anomaly detection algorithms have leveraged different tools, for example, nearest-neighborhood and clustering-based or spectral-based anomaly detection algorithms. Some of the commonly used machine learning anomaly detection algorithms include self-organizing maps (SOM) [47], K-mean [39], adaptive resonance theory (ART) [48], and one-class support vector machine (OC-SVM) [49]. An excellent survey of anomaly detection can be found in Ref. [44]. We will provide more details and examine the use of unsupervised learning in reliability engineering and safety application in section 3.

## 2.3 Semi-supervised learning (SSL)

In some problems, such as image recognition, photo categorization, and autonomous driving, there can be a limited labeled dataset, i.e., paired observations ($X_i$,$Y_i$), along with a much larger unlabeled dataset of features ($X_j$ with no $Y_j$). Obtaining unlabeled data is often easy or cheap, whereas expanding the labeled dataset can be time-consuming or expensive. In these situations, neither supervised nor unsupervised are suitable for handling these problems. Semi-supervised learning is a set of methods for these types of problems, and which bridge the gap between supervised and unsupervised approaches. They use the labeled dataset to estimate $\hat{f}$ and leverage the much larger unlabeled dataset, which under certain assumptions[2] can help improve the model's performance and provide more accurate predictions, as shown in Fig. 4 in the context of a classification problem. Semi-supervised learning methods have been used in speech recognition, internet content classification, bioinformatic [50], and a host of other applications. These methods leverage, for example, generative models [51] and graph-based algorithms [52] to learn from the mixed labeled–unlabeled data (they differ by how the unlabeled data is used, and by the choice of the loss function and regularizer). Two excellent surveys of semi-supervised learning can be found in Ref. [53] and the more recent Ref. [54]. We will provide more details and examine the use of semi-supervised learning in reliability engineering and safety application in section 3.

---

[2] Generally described as the smoothness assumption, the cluster assumption, and the manifold assumption for classification problems. These assumptions are related, for example the smoothness assumption states that if two points are in a high-density region, then so should their corresponding labels. The cluster assumption states that if points are in the same cluster (unsupervised learning), then they are likely to be in the same class as well (supervised learning). This is also known as the low-density separation assumption, which states that the decision boundary should lie in a low-density region [53, 54]. See Fig. 4 for an illustration.

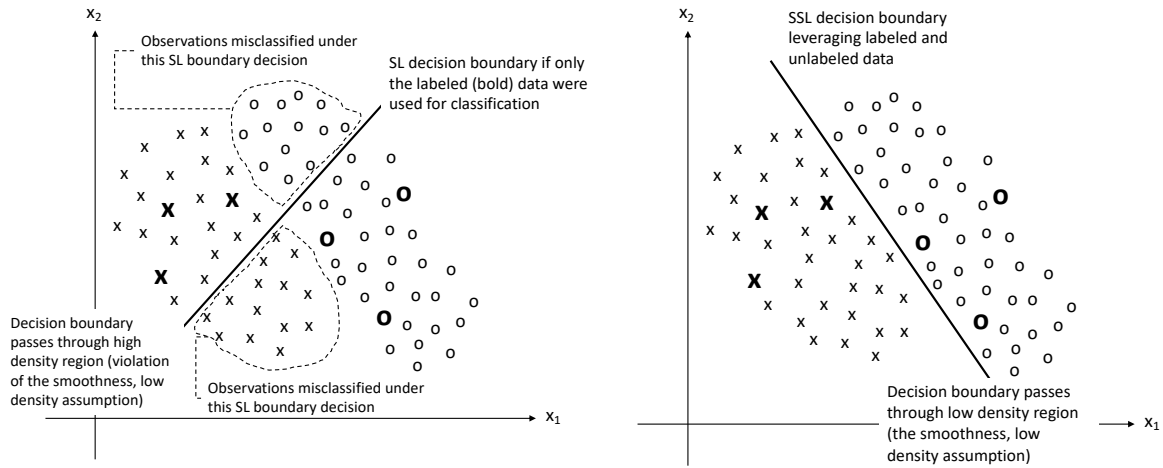

**Figure 4**. Illustration of classification under supervised (left panel) and semi-supervised learning. The 'x' and 'o' belong to two separate classes, but only the bold ones are labeled. If the learning algorithm only makes use of the labeled data (supervised mode), the decision boundary will pass through high-density regions in the feature space and misclassify many data points (poor accuracy). A semi-supervised learning algorithm will leverage the low-density, cluster assumption to produce the decision boundary in the right panel, which improves the classification accuracy. Adapted from Ref. [53].

## 2.4 Reinforcement learning (RL)

Reinforcement learning is a distinctive set of techniques, which, unlike the previous ML categories, does not experience or operate with a fixed dataset. Instead, reinforcement learning algorithms interact with an environment, and as a result, their experience becomes part of the dataset from which they learn. They explore by trial-and-error interactions with the environment during the training process. A pre-requisite for this feedback loop between experience and learning is the presence of a reward function. Reinforcement learning is therefore used in different contexts, by software and machines, to make decisions (sequentially) and find a course of action (policy) that maximizes the agent's cumulative reward. The reward can be defined for a specific outcome or with a specific function of the state of the agent. A fundamental issue in reinforcement learning algorithms is setting parameters that achieve a proper balance between agent exploration and exploitation to balance the stability and efficiency of the learning process [55]. An over-exploration agent may have difficulties converging to an optimal course of actions, whereas an over-exploitation agent is likely to get trapped in a local optimum and fail to find the global optimal solution. Reinforcement learning has very broad applications, including robotics and autonomy [56], gaming [57], and online advertisement [58]. The major reinforcement learning algorithm includes policy gradient [59], Q-learning [60], deep Q network (DQN) [61], and actor-critic algorithm [62]. We will examine the use of reinforcement learning in reliability engineering and safety application in the next section.

## 3. Machine learning applications in reliability engineering and safety applications: overview and current status

In this section, we review the current use of ML in reliability engineering and safety applications. We also introduce some of the algorithms and provide a short guide to the growing literature in these areas. While no claim of exhaustiveness can be made when undertaking such a task, we have attempted to sample major publications in each category and subcategories of ML discussed previously. In addition, we devote the last subsection to the use of Deep Learning (DL) in reliability and safety applications, as separate from the rest despite its overlap with the previous categories to highlight its growing popularity and emphasize its distinctive advantages over other types of ML.

The basic ingredients for a machine learning algorithm are the following: (i) datasets for training and testing purposes, (ii) an objective function or loss function to optimize, for example, a sum of squared errors or a likelihood function, and (iii) an optimization algorithm and a model for the data (e.g., linear, nonlinear, nonparametric). By varying any one of these ingredients, it is easy to conceive of the very wide range of applications of ML for reliability and safety applications, for example by examining different datasets in various industries, by applying different ML models to various systems or components, and by modifying some models and algorithms to better suit the task at hand. The objective is generally to tease out novel, more accurate results from datasets for better reliability and safety-informed decision-making in system design and operation, and more effective accident prevention. Some of the recurrent themes in the literature we examine next include ML use for the estimation of an asset remaining useful life, anomaly and fault detection, health monitoring, maintenance planning, and degradation assessment.

### 3.1 Supervised learning applications

In this subsection, we review some of the regression and classification applications, and we provide brief introductions to some of the corresponding ML models.

**3.1.1 Regression**

Supervised regression is widely used for remaining useful life (RUL) estimations and degradation predictions. The literature in this area includes applications of ML to different engineering items, for example, Lion-ion batteries [63], railway tracks [64], turbine cutting tools [65], rolling bearings [66], and aircraft engine [67, 68]. In addition to these different domains of application, researchers have explored different ML models, for example, support vector machine/regression [67], Gaussian Process Regression (GPR) or kriging [69], and Deep Learning for structural reliability problems, RUL prediction, fire hazard simulation, and more generally for engineering surrogate modeling applications [70]. A selection from this ML regression literature is provided in Table 1.

Table 1. Selection from the literature on ML regression models in reliability and safety applications

| ML model | Reported advantage | Application | Ref. |
| --- | --- | --- | --- |
| **Vector machine:** | | | |
| Support vector regression (SVR) | 1.Superior prediction accuracy than the traditional method<br>2.Robust to data noise<br>3.Superior efficiency than traditional simulation models. | 1.Failure and reliability prediction of time series data<br>2.RUL estimation<br>3.RUL estimation for aircraft engines<br>4.Approximations of fire hazard model at nuclear power plants | [65]<br>[67]<br>[71]<br>[72] |
| relevance vector machine (RVM) | 1.Superior prognostic accuracy than traditional methods | 1.System degradation prognostic | [63] |
| **Neural network:** | | | |
| Deep neural network (DNN) | 1.Excellent prediction accuracy and training efficiency<br>2.Excellent long- and mid-term prediction accuracy | 1.RUL of aircraft engine prediction<br>2.Human errors prediction<br>3.Component reliability and degradation level prediction | [64]<br>[68]<br>[73] |
| Convolutional neural network (CNN) | 1.Excellent prediction accuracy to highly nonlinear, complex, multidimensional system | 1.RUL estimation | [74] |
| Recurrent neural network (RNN) | 1.No need for prior knowledge and analysis of the dataset | 1.RUL estimation | [75] |
| Deep bidirectional long short-term memory (LSTM) | 1.No need for prior knowledge and assumptions<br>2.More accurate estimation than traditional methods | 1.RUL estimation | [76] |
| CNN based LSTM | 1.Suitable for more complex modern engineering system, and high dimensional input | 1.Multi-scale feature selection and RUL estimation | [66] |
| LSTM and gated recurrent unit | 1.Excellent efficiency and accuracy for nonlinear complex system | 1.RUL prediction and PHM | [77] |
| **Gaussian process:** | | | |
| Gaussian process regression (GPR) | 1.Superior accuracy and efficiency than traditional methods<br>2.Suitable for nonlinear high dimensional system analysis<br>3.Dynamic updating of the model parameter<br>4.Active learning and excellent performance in complicated real-world applications | 1.Time-dependent failure probability prediction<br>2.Structural reliability analysis and failure probability estimation<br>3.System reliability analysis<br>4.Failure-pursuing sampling framework<br>5.Reliability-based importance analysis of structural system | [69]<br>[78]<br>[79]<br>[80]<br>[81]<br>[82] |

An accurate prediction of an equipment RUL and degradation level is essential for a safety-critical system, and it is important for condition-based maintenance to minimize system downtime and other adverse consequences of a run-to-failure approach. RUL prediction is an important element of prognostics and health management (PHM), and the demand for better PHM requires, among other things, a more accurate estimation of the RUL. Approaches to RUL and degradation level predictions can be grouped into two broad categories, with a third hybrid one straddling the first two: (i) model-based approaches, which build failure models based on the detailed analysis of the physical nature of the failure mechanism under consideration [83, 84]. These models require extensive prior knowledge and subject-matter expertise; (ii) data-driven approaches, which build degradation models from historical sensor data, and therefore require no prior knowledge of the system. These models can be built with various ML approaches, and they can vary in accuracy and computational intensity given the quality and quantity of data available. We briefly introduce next two powerful and commonly used ML regression models in reliability and safety application, namely support vector regression (SVR) and Gaussian Process Regression (GPR). Deep Learning regression models are discussed in subsection 3.5.

Support vector machine (SVM), also known as Support Vector Regression (SVR) when applied to regression problems, is a supervised ML approach that can be used for both regression and classification tasks. The principle of SVM is to construct a set of hyperplanes in the feature space ($X_i$) to conduct regression or classification with the input data. The parameters are calculated such that the hyperplanes are at a maximum distance from the nearest training data points in order to reduce the classification or regression error. The original SVM model was designed to handle problems that are linearly separable in the design space (features). However, real-world applications often exhibit nonlinearities in their features ($X_i$) and their relationship to the output variable ($Y_i$). To overcome this drawback, the kernel method is used with SVM, which maps the original low dimensional non-linear problem into a higher dimensional space where the data is linearly separable. For example, the Gaussian radial basis function, which maps the original data to infinite-dimensional space [22], is a popular kernel for SVM applications. The reader is referred to Ref. [22] for more details on SVM method. Variants of SVM, such as the relevance vector machine (RVM) [85] have also been developed for equipment degradation and RUL prediction applications.

GPR, also known as Kriging, is a nonparametric Bayesian statistical learning approach to regression problems. In GPR, the response function is defined as a stochastic posterior distribution in the inference equation, with the Gaussian process prior conditioned on the input dataset. The GPR infers a probability distribution with mean and standard deviation estimations of the normally distributed response function. With a proper selection of Gaussian process kernel and corresponding hyperparameters, GPR and its variants such as active learning Kriging (ALK), can provide accurate response function estimation and uncertainty quantification [86], and they are increasingly used in engineering surrogate modeling and reliability analysis, [87, 88]. The reader is referred to Ref. [89] for more details on the GPR.

**3.1.2 Classification**

Supervised classification is widely used in fault detection and identification. This includes both online monitoring of the degradation states of equipment and diagnosing the type of faults that occur (binary and multi-class). Classification in this context is at the nexus of two broader considerations, PHM and predictive maintenance, and it provides critical information that helps inform the conduct of both. The literature in this area includes the use of classification ML tools for fault detection and identification in a wide range of reliability and safety applications, for example, aircraft engine and electric power transformers [90], water distribution and pipe failures [91], bearings or rotary machines [92, 93], wind turbine blades [94], software reliability [95], and forest fires [96]. In addition to these different domains of application, researchers have also developed or adapted ML classifiers for specific PHM purposes. For example, Islam and Kim [92] developed an improved one-against-all multiclass support vector machine classifier and used it with acoustic signals to diagnose eight types of fault classes in bearings (seven different fault types and one nominal operational conditions). Stern et al. [97] used ML classifiers (SVM and logistic regression) to develop surrogate models for the reliability of a complex two-terminal network. They demonstrated high prediction accuracy and order of magnitude smaller computational effort than traditional Monte Carlo Simulation. Rachman and Ratnayake [98] benchmarked the performance of different ML classifiers for equipment screening and risk-based inspection assessment (with application to oil and gas pressure vessels and piping lines). They demonstrated better accuracy and precision with ML classifiers than conventional methods of screening for inspection and a reduction of output variability between human appraisers. A selection from this ML classification literature is provided in Table 2.

Table 2. Selection from the literature on ML classification models in reliability and safety applications

| Technique | Reported advantage | Application | Ref. |
|---|---|---|---|
| **Conditional models:** | | | |
| Decision tree (DT) | 1.Excellent accuracy with great training efficiency | 1.Assessing stakeholders corporate governance<br>2.Dirt and mud detection on a wind turbine blade<br>3.Forest fire risk assessment | [94]<br>[96]<br>[99] |
| Random Forest (RF) | 1.Suitable for discrete classification<br>2.Excellent estimation accuracy | 1.Rank the importance of each component an engineering system | [100] |
| Boosted Tree | 1.Excellent accuracy and overfitting free | 1.Risk-based inspection screening assessment | [98] |
| **Nearest neighborhood** | | | |
| K-nearest neighborhood (KNN) | 1.Excellent accuracy and efficiency | 1.Risk-based inspection screening assessment<br>2.Dirt and mud detection on a wind turbine blade | [94]<br>[98] |
| **Support vector:** | | | |
| Support vector classification (SVC) | 1.Highly efficient with up to two orders of magnitude time saving compared with traditional methods<br>2.Robust to appraiser-to-appraiser output variation, and better prediction accuracy<br>3.Robust to dataset noise | 1.Risk-based inspection screening assessment<br>2.Dirt and mud detection on a wind turbine blade<br>3.Bearings fault diagnosis<br>4.Early detection of gradual concept drifts<br>4.Reliability analysis of network connectivity | [92]<br>[94]<br>[97]<br>[98]<br>[101] |
| **Bayesian method:** | | | |
| Linear discriminant analysis (LDA) | 1.Excellent prediction accuracy | 1.Dirt and mud detection on a wind turbine blade | [94] |
| Bayesian net | 1.Available for a complex system with excellent accuracy and efficiency<br>2.Suitable for discrete classification and satisfying estimation | 1.Water distribution system pipe failures analysis<br>2.Rank the importance of components of the engineering system<br>3.Seismic loss analysis of spatially distributed infrastructure system | [100]<br>[91] |
| Gaussian process classification (GPC) | 1.Suitable for complex system<br>2.Excellent computational efficiency | 1.Reliability evaluation of complex system | [102] |
| **Neural network:** | | | |
| Stacked autoencoder | 1.Robust to the input noise<br>2.Excellent accuracy | 1.Bearing fault diagnosis | [93] |
| Recurrent neural | 1.Superior prediction accuracy than | 1.Software reliability prediction | [95] |

| network (RNN) | traditional parametric models | | |
|---|---|---|---|
| Deep belief network | 1.Efficient inference | 1.Health state classification | [90] |
| | 2.Able to encode richer and higher-order complex systems | | |
| Long short-term memory (LSTM) | 1.Capable of the safety analysis of time-varying systems | 1.Dynamic predictive maintenance framework for failure prognostics | [103] |
| | 2.No need for prior assumption and knowledge | | |

ML classification is enabled by a set of models, from the simple k-nearest neighbor (KNN) and logistic regression to more advanced Decision Trees (DT), linear discriminant analysis (LDA), and support vector classification (SVC). In addition to these single-classifier methods, ensemble classifiers have been developed that incorporate multiple single classifiers for better performance. These include, for example, Random Forest (RF), and AdaBoost (AB). As with regression models, the availability of a wide range of ML classifiers makes it important for research in reliability and safety applications to benchmark and assess the performance of different classifiers, and then select the most appropriate one given the datasets used. This model selection phase makes a work more convincing and lends its findings more weight. Unfortunately, it is missing in some publications, and we recommend that authors, reviewers, and editors be more mindful of this expectation of transparency for comparative model analysis and selection when writing or reviewing manuscripts in this area of applications.

We briefly introduce next one popular ML classifier in reliability and safety application, namely decision tree (DT) to illustrate some aspects of ML classification. DT uses a tree-like structure in which each node leads to different branches based on a threshold value for a covariate or feature variable [104]. In essence, DT slices the design space or feature space into multiple regions. Then, for predicting the response variable of a new test case, the DT classifier identifies the feature region in which the new case resides. For this reason, DTs are described as conditional control-based algorithms in which each node of the tree structure represents a "test" on a feature, and each branch a possible outcome of the test [104]. In order to improve the accuracy and avoid potential overfitting with DT, ensemble DTs such as random forest (RF), have been developed. The DT and its variants are suitable for classification problems where the decision boundaries between different classes are sharp. Other classifiers may be better suited for handling smooth boundaries, for example, support vector classification (SVC) and Bayesian classification models which are also popular in fault detection and identification and operate with somewhat similar mechanisms to SVR and GPR regression models discussed previously. More details can be found in Ref. [35].

## 3.2 Unsupervised learning applications

In this subsection, we review some of the unsupervised clustering and anomaly detection applications, and we provide brief introductions to some of the corresponding ML models.

### 3.2.1 Clustering

Clustering or unsupervised classification is not as widely used in reliability and safety applications as its supervised counterpart. It has significant untapped potential, which other fields mentioned previously, such as image recognition, genomics, and e-commerce leverage to a fuller extent. Some applications of clustering include degradation analysis in railway point machines [105], wind turbine failures [106], fault detection in the nuclear industry [107], damage classification in structural components [108], bearing faults identification in rotating machines [109], and blasting operations in mining activities [110].

Beyond these specific areas of applications, some authors have also leveraged clustering tools for broader, methodological problems in reliability and safety. For example, Fang and Zio [111] adapted a form clustering for network reliability and criticality analysis. They leveraged clustering to demonstrate a novel hierarchical modeling framework and to extract structural properties of the network (in a reliability sense). Soualhi et al. [112] developed a novel unsupervised classification technique called Artificial Ant Clustering (AAC). They successfully applied it to identify operational modes and diagnose two types of faults in an induction motor (broken rotor bars and bearing failures). The authors demonstrated significant improvements in clustering error rates compared with other methods for their particular application, even when information about the different operating modes of the motors was limited. A selection from this ML clustering literature is provided in Table 3.

**Table 3.** Selection from the literature on ML clustering (unsupervised classification) models in reliability and safety applications

| Technique | Reported advantage | Application | Ref. |
|---|---|---|---|
| **Self-organizing map:** | | | |
| Self-organizing map (SOM) | 1. Capable of detecting various levels of fault with high accuracy  2. Promising result in early fault detection of real application with excellent visualization analysis | 1. Fault detection and isolation scheme for pneumatic actuator  2. Early Fault detection of the nuclear industry | [105] [107] [113] |
| Principle component analysis-self-organizing map (PCA-SOM) | 1. High efficiency to high dimension system with excellent accuracy | 1. Damage classification in structural health monitoring | [108] |
| **Distance metric learning:** | | | |
| K-mean | 1. High detection accuracy and efficiency | 1. Associating weather condition and wind turbine failures | [106] |
| Neighborhood component analysis | 1. Efficient for high dimensional input data with outstanding accuracy | 1. Fault detection to bearings | [109] |
| Spectral clustering | 1. Capable of identifying the most relevant clusters | 1. Network component-level safety and support criticality analysis | [111] |
| **Bayesian method:** | | | |
| Bayesian net | 1. Accurate prediction with limited information | 1. Analyze the risk of accident in complex blasting operations | [110] |
| **Deep neural network:** | | | |
| Long short-term memory (LSTM) | 1. Excellent efficiency and accuracy | 1. Estimate the health index of a system | [114] |

ML clustering is enabled by a set of models noted previously such as the popular K-means algorithm and principal component analysis (PCA), as well as a wide range of hierarchical clustering methods. We briefly introduce next two popular clustering models in reliability and other applications, namely the self-organizing map (SOM) and the distance metric learning (DLM). SOM is an unsupervised learning method based on neural networks. It is a popular dimension reduction tool that maps the high-dimensional feature space to a smaller, typically two-dimensional space, and it produces a low dimensional representation of the input data. Following this dimensionality reduction, SOM identifies clusters in the new feature space based on some similarity measures such as Euclidean distance [115, 116] Details about SOM can be found in Ref. [117]. Examples of the use of SOM in reliability and safety applications are provided in Table 3.

Distance metric learning is a category of machine learning algorithms that extracts similarity information from the input features themselves. It can be applied with both supervised classification tools such as KNN, and unsupervised clustering tools such as K-mean. Several ML algorithms rely on being provided a good similarity metric over their input data to perform effectively. The quality of their output is highly dependent on the similarity metric used, and DLM can considerably improve their classification or clustering performance. Details about distance metric learning can be found in Ref. [118, 119]. Examples of the use of Distance metric learning in reliability and safety applications are provided in Table 3.

### 3.2.2 Anomaly detection

Anomaly detection is particularly well suited for and used in early fault/damage detection of engineering equipment and structures. It is intrinsically related to sensor data, and for industrial machinery and equipment, the data typically comes in streaming fashion (temporal dimension). Early detection of anomalies is essential in some contexts to prevent further damages and preempt catastrophic failures. For structures such as beams, airframes, or bridges, the data for anomaly detection has both spatial and temporal dimensions. Although we have listed anomaly detection under unsupervised learning, and we discuss here under this heading, algorithms for anomaly detection also exist in supervised and semi-supervised mode. That being said, the more widely used though are in unsupervised mode, often because unlabeled data is widely available and labeled data is expensive (and often rare) to obtain. In reliability and safety applications, data is often available for (labeled) nominal operational conditions only. When that is the case, anomaly detection can leverage tools of semi-supervised learning since partial labels are available.

The literature includes applications of anomaly detection in support of PHM for different engineering systems, for example, aircraft flight data recorders [120], industrial gas turbines [121], spacecraft operation and health monitoring [122, 123], and induction motors with a focus on ball-bearing faults [124]. Applications of anomaly detection algorithms for structural damage detection also abound, a discussion of which can be found in [44]. A selection from this ML anomaly detection literature is provided in Table 4.

Many of the same tools of clustering are also used or adapted for anomaly detection. Some of the recent works on anomaly detection in reliability and safety applications also made important methodological contributions to the field. For example, Ince et al. [124] leveraged Convolutional Neural Networks (CNN) to develop a highly accurate and robust detection method of faults in bearings of induction motors. Feature extraction is typically computationally expensive, and the accuracy of anomaly detection algorithms often hinges on the careful selection of features. The authors addressed both limitations by developing a novel CNN model that merges feature extraction and (unsupervised) classification into a single learner. They validated their method by directly using raw sensor data to detect anomalies (no preprocessing or input transformation). This is a meaningful contribution that can enable real-time anomaly detection capability with streaming data. Yan and Yu [121] used Deep Learning tools for detecting abnormal behavior and incipient faults in industrial turbines where failures can be catastrophic and extremely costly. The authors recognized that "advanced technologies that can improve detection performance are in great need" in their domain of application. Their model architecture integrated a stacked denoising autoencoder [125] with a neural network known as Extreme Learning Machine (ELM) to significantly improve on the sensitivity and specificity of other anomaly detection methods for gas turbines. The SDAE+ELM model uses exhaust gas temperature measurement to infer the operational condition of the combustor in the gas turbine and detect anomalies. Just like Ref. [124], Yan and Yu [121] demonstrated that their ML model can automatically extract the right features—a critical and challenging task—from the raw time-series temperature measurements to improve the anomaly detection without the time-consuming "handcrafted" feature extraction phase.

Table 4. Selection from the literature on ML anomaly detection models in reliability and safety applications

| Technique | Reported advantage | Application | Ref. |
| --- | --- | --- | --- |
| **Support vector:** | | | |
| Support vector machine (SVM) | 1.Excellent accuracy and efficiency for feature extraction 2.Accurate in detecting the early signs of system anomalies | 1.Real-time Motor machine failure identification and early fault diagnosis 2.Spacecraft health monitoring | [122] [124] |
| **Neural net:** | | | |
| Convolutional neural network (CNN) | 1.Excellent accuracy and efficiency for feature extraction | 1.Real-time Motor machine failure identification and early fault diagnosis | [124] |
| Deep neural network (DNN) | 1.General for different applications and prior assumption free. | 1.Anomaly detection for gas turbine combustors | [121] |
| Deep auto-encoder | 1.Accurate and efficient fault detection in aircraft flight data | 1.Anomaly detection and fault disambiguation in large flight data | [120] |
| Long short-term memory (LSTM) | 1.Capable of supervising large amount of high dimensional spacecraft telemetry data with good accuracy 2.Capable of performing anomaly detection without costly expert knowledge | 1.Detecting spacecraft anomalies | [123] |

One challenging domain of application is worth discussing as the literature in this area will likely witness a significant growth in the coming years, namely anomaly detection for spacecraft operation and health monitoring. Spacecraft can be particularly complex and expensive machines with thousands of telemetry data points for monitoring the health and performance of all its subsystems and payload. Since physical access to the craft is limited or impossible, detecting anomalous behavior as early as possible is particularly important in order to provide satellite operators with sufficient lead-time to prevent the incipient fault for further developing, or to develop a workaround that limits its damage and consequences. Some spacecraft missions can downlink gigabytes to terabytes of data per day, and as a result, there is a critical need for: (i) automated anomaly detection methods that can support/augment limited engineering flight resources for manual monitoring of spacecraft health; (ii) highly accurate and robust detection algorithms (high sensitivity and specificity), since on the one hand a missed detection can lead to dramatic consequences such as complete loss of mission, and on the other hand given the high volume of telemetry data provided, even a small false alarm rate will lead to a voluminous number of false alarms and overwhelm the operators (thus defeating the purpose for which the anomaly detection system was designed in the first place, to alleviate the operators' workload); and finally (iii) highly scalable anomaly detection algorithms that can operate efficiently with large multivariate streaming data. These challenges are far from being resolved. They are likely to become more critical in the near future, and there are significant opportunities for progress and meaningful contributions in this domain.

To date, spacecraft anomaly detection is mostly based on fixed-threshold alarms or out-of-limit (OOL) method, thus treating the problem as a detection of *point anomalies*. This is appropriate in some cases, but it misses the richness of the context and the signatures of many other *contextual anomalies* in times series data for early detection. A more insidious limitation of this approach is that it assumes that all anomalous behaviors are known beforehand, and the corresponding thresholds are set and programmed to be detected. This of course is rarely the case, and as a result, current spacecraft anomaly detection systems only target a small subset of possible anomalies, and those that are likely to be detected are not flagged early enough (or at their earliest possible detection). Two recent contributions that tackled these problems and worth noting. Fuertes et al. [122] proposed a One-Class Support Vector Machine (OC-SVM) anomaly detection algorithm with some pre-processing of the input telemetry data. Their algorithm showed promising results and detected anomalies that had not been detected by current monitoring systems. But it came at the price of a high false alarm rate, and the authors recognized that this challenge should be addressed "before this type of surveillance can be accepted and trusted by [spacecraft] operational teams." Hundman et al. [123] examined (expert-labeled) anomalies from the Soil Moisture Active Passive (SMAP) satellite and the Mars Science Laboratory (MSL) Curiosity rover, and they reported that 41% were *context anomalies*. Thus, a significant proportion of spacecraft anomalies are not suitable for or cannot be handled effectively with the prevalent OOL approaches and detection algorithms for point anomalies. The authors then leveraged advances in Deep Learning [17] to develop a type of recurrent neural network (RNN) known as Long Short-Term Memory (LSTM) for spacecraft anomaly detection that is capable of flagging context anomalies. LSTM is well-suited for modeling temporal data and prioritizing historical information for future predictions [126, 127]. The authors demonstrated the viability of LSTM for detecting anomalies in near real-time spacecraft telemetry with over 700 channels. However, one major obstacle for deployment they reported was the same as the one faced by Fuertes et al. [122], namely high rates of false positives. "High demands are placed on operations engineers, and they are hesitant to alter [existing] procedures. Adopting new technologies [for spacecraft anomaly detection] means an increased risk of wasting valuable time and attention. Investigation of even a couple of false alarms can deter users, and therefore achieving high precision […] is essential for adoption."

In short, many challenges remain but also significant opportunities exist for leveraging anomaly detection algorithms (and more generally unsupervised and semi-supervised learning) in a host of reliability and safety applications. We will discuss some of them in section 4.

## 3.3 Semi-supervised learning applications

Semi-supervised learning algorithms are used in fault detection and identification, in prognostics and RUL prediction, which are important for maintenance planning. These methods have significant, and in our opinion, barely tapped potential for reliability and safety applications where unlabeled data is abundant but labeled (e.g., failure) data is scarce and expensive to come by, or it takes a long time to collect. The literature in this area includes applications of SSL in support of fault detection and PHM or RUL prediction for different engineering systems, for example cooling fans [128], centrifugal pumps [129], turbofan engines [130, 131], bearing defects in induction motors [132], and solar arrays [133]. Hu, Youn, and Kim [128] used a co-training regression SSL algorithm for the RUL prediction of bearing and cooling fans. They demonstrated better accuracy in RUL prediction, along with a smaller scatter of the results, compared with a SL learning algorithm. They used nonetheless a high ratio of labeled to total data (50%), which is much larger than what is typically found in SSL applications. He et al. [129] used ladder network (LN), a form of deep denoising auto-encoder to estimate the RUL of centrifugal pumps, with an unsupervised feature extraction operating offline, and the prediction performed online with streaming data. Yoon et al. [131] used a different strategy to estimate the same quantity, the RUL of turbofan engines, but using a variational auto-encoder (VAE) trained on both the unlabeled and labeled data simultaneously, and with a fraction of the labels down to 1% of the entire dataset. Both works demonstrated good prediction accuracy even when the available label information was highly limited. A selection from this SSL literature is provided in Table 5.

**Table 5.** Selection from the literature on semi-supervised learning in reliability and safety applications

| Technique | Advantage | Application | Ref. |
|---|---|---|---|
| **Graph based model:** | | | |
| Graph based model (GBM) | 1.Excellent self-learning ability  2.Excellent data visualization  3.Excellent accuracy and great efficiency | 1.Fault detection  2.Health prognostic and maintenance activity identification | [133]  [134] |
| **Support vector machine:** | | | |
| One class support vector machine (OCSVM) | 1.Excenlent near miss prediction | 1.Near-miss fall detection | [135] |
| Semi-supervised support vector machine (S3VM) | 1.Accurate for complex high-dimensional system | 1.Fault diagnostic | [132] |
| **Neural network:** | | | |
| Deep generative models | 1.Superior accuracy than traditional methods | 1.Asset failure prediction | [131] |
| Ladder network | 1.Superior efficiency and accuracy than traditional methods | 1.RUL prediction for centrifugal pumps | [129] |
| Long short-term memory-restricted Boltzmann machine (LSTM-RBM) | 1.Reliable RUL prediction with insufficient labeled data | 1.RUL prediction for turbofan engine | [130] |
| Radial basis network (RBN) | 1.Excellent accuracy and robustness. | 1.Data-driven accident prognostics, RUL prediction | [128] |

Co-training is an example of "wrapper methods" in SSL, which are simple approaches to extending supervised algorithms to the semi-supervised mode by performing the following steps: first, a regressor or classifier is trained on the labeled data, and predictions are performed on the unlabeled data. Second, the most confident predictions are pseudo-labeled and added to the training dataset. The regressor or classifier is then retrained on the expanded dataset in a supervised mode, "unaware of the distinction between originally labeled and pseudo-labeled data" [54]. The process is repeated until a termination criterion is reached and the regressor or classifier cannot be further improved.

Although we mentioned previously examples of semi-supervised regression problems, the majority of applications of semi-supervised learning are in classification, and in reliability and safety applications, these revolve around fault detection and identification. We briefly discuss next one interesting work in this area. Zhao et al. [133] developed a semi-supervised method for fault detection and classification in solar photovoltaic (PV) arrays. The authors noted that "without proper fault detection, unnoticed faults in PV arrays might lead to safety issues and fire hazards. Conventional fault detection and protection usually add overcurrent protection devices (OCPD)", but certain faults have been shown to be unnoticeable by these devices. Furthermore, fault classification (identification of the types of fault) can be particularly helpful in helping to identify the maintenance needs to expedite the recovery of the system. The authors developed a graph-based method (GMB) that uses existing measurements such as PV voltage, current, and operating temperature (no additional hardware required than currently installed), with a low training cost leveraging a large pool of unlabeled data (less than 2% of the data is labeled) and self-learning to achieve a remarkable detection and classification accuracy (99%) under real working conditions. GBMs are the focus of one of the most active areas of research in semi-supervised learning. Their common denominator is that the data are represented by the nodes ($V$) of the graph, and the edges ($E$) are labeled by a measure of similarity between two nodes, the Euclidean distance for example or some other (geodesic) distance. GBMs are based on the manifold assumption, which roughly states that high dimensional data lie on a low dimensional manifold, and nearby data in the feature space have similar label prediction. In most instances, the predictions of GBM consist in labeling the unlabeled dataset based on local similarities between nodes or a weighted distance of the edges. "A parallel can be drawn between [graph-based] methods and supervised nearest-neighbor methods. The latter predicts the label of an unlabeled data point by looking at the labels of similar (i.e., nearby) labeled data-points; graph-based methods also consider the similarity between pair of unlabeled data points. Using that information, labels can be propagated transitively form a labeled data point to an unlabeled data point" [54]. The reader is referred to Ref. [136] for more details on GBM. Other approaches to SSL include SVM-based methods such as semi-supervised support vector machine (S3VM) and one-class support vector machine (OC-SVM) [137, 138].

One last note of caution regarding SSL methods the reader should be aware of the publication bias and the importance of safe semi-supervised learning. While the vast majority of articles on SSL report improved accuracy and better prediction than their SL counterparts, this needs not to be the case in real-world applications. The performance of a regressor or classifier can degrade when unlabeled data is exploited alongside labeled data if the underlying assumptions of SSL are violated. Such findings would rarely be reported given the publication bias toward positive results, thus skewing the perception that SSL methods are unconditionally successful. The theme of *safe semi-supervised learning* is a growing research area with meaningful recent progress, and it examines the circumstances under which unlabeled data can confidently be integrated into the learner its without degrading its performance[3]. The reader is referred to Li and Liang [139] and the references therein for an introduction to these issues.

## 3.4 Reinforcement learning applications

Reinforcement learning is not yet as broadly used in reliability and safety applications as supervised or unsupervised learning. Yet, it offers significant opportunities for making important contributions to these areas. We briefly review next four recent publications from 2019 and 2020.

Some recent applications include reliable handover in cellular network operations for mobility robustness optimization (MRO). This is a challenging problem for traditional rule-based methods, the objectives of which are to minimize the number of dropped calls/unsatisfied customers, increase each cell throughput, and ensure a more balanced network using cell load-sharing [140]. The authors developed a Deep Reinforcement Learning (DRL) solution that outperformed (on user QoS) and required fewer parameters to tune than traditional methods for reliably handling wireless user handover across cells. Reinforcement learning has also been used for condition-based maintenance planning with multi-component systems subject to competing risks [141]. Condition-based maintenance (CBM) has traditionally been examined with Markov Decision Processes (MDP) [142] and its variant the partially observable Markov decision processes (POMDP) [143, 144, 145]. MDP and POMDP can efficiently handle small systems with a limited number of states and possible actions for decision-making. They strain, however, at handling large (real) systems in which the number of states and actions grows exponentially with the number of components. For example, a system with "20 components, 5 states and 5 actions per component is described by nearly $10^{14}$ states and actions! This renders the problem practically intractable by any conventional scheme or advanced MDP or POMDP algorithm" [146]. With the increasing complexity of engineering systems, optimization of CBM becomes challenging if not computationally intractable with MDP and other conventional decision-making tools. Andriotis and Papakonstantinou [146] leveraged Deep Reinforcement Learning to identify efficient inspection and maintenance policies for large-scale infrastructure systems. Xiang et al. [147]

---

[3] *Safe* in this context is not used in its traditional sense of free from danger or harm, but to "indicate that the performance [of the SSL learner] is never worse than methods using only labeled data" [139]. The corresponding noun in this context is *safeness*, not *safety*.

developed a Deep Reinforcement Learning-based sampling method for structural reliability assessment. The sampling space is treated as the state of the DRL method, and the sample selection as its action (coupled with a reward function to guide the deep neural network in selecting the sampling points). The proposed method achieved a higher prediction accuracy of failure probability than competing sampling methods such as Monte Carlo Simulation and Latin hypercube sampling. Reinforcement Learning techniques are well suited to alleviate the curse of dimensionality in large state and action space systems, and as such they are ideal for handling the scheduling of inspection and maintenance of large engineering systems across their life cycle. The recent advances in Deep Reinforcement Learning [61] make these tools even more effective at identifying near-optimal policies for traditionally intractable systems and other computationally intractable problems in reliability and safety applications.

One last important subfield of research within RL is worth mentioning as the literature and applications in this area are likely to experience significant growth in the coming years, that is, Safe Reinforcement Learning (SRL). In all the previous discussions, we noted how ML algorithms have been used to address a reliability or safety application; in SRL the reverse relationship holds, namely safety considerations are brought to bear on the learning algorithm to avoid exploring hazardous or unsafe states (known more generally as *error states*) during the learning process. SRL is defined as "the process of learning policies that maximize the expectation of return in problems in which it is important to respect safety constraints" during the learning and deployment phases[4] [148]. This is crucial in some situations, for example when the learning phase occurs online with hardware in the loop, not through simulators. Reinforcement Learning, as noted in subsection 2.4, involves an agent perceiving the state of the environment and learning a course of action (policy) that maximizes its cumulative reward. In many robotics applications, whether the algorithm is learning to fly a drone or perform surgery for example, safety is particularly important, and avoiding damage to the agent or the environment is essential (e.g., it is not an efficient learning algorithm if it crashes a drone repeatedly before it is capable of skillfully flying it). Safety constraints or preferences have been incorporated into RL algorithms either by modifying the optimization criterion, or by modifying the exploration process through the use of risk metrics, or by incorporating external knowledge about undesirable states [148]. With the proliferation of social robots, Safe Reinforcement Learning will remain a vital research area, and its applications will extend to incorporate different forms of risks. For more details on Safe Reinforcement Learning, the reader is referred to Garcia and Fernandez [148] and Munos et al. [149] and the references therein.

---

[4] There are other definitions of safe reinforcement learning, see for example Munos et al. [149]

## 3.5 A note on deep learning

This last subsection overlaps with the previous ones since in discussing earlier applications leveraging for example CNN or LSTM, we were in fact presenting some of the use of deep learning (DL) in reliability and safety applications. We include this subsection nonetheless to highlight DL's growing popularity and emphasize its distinctive advantages over other types of ML. DL is a subset of machine learning ($DL \subset ML$), and it is capable of handling all categories of ML shown in Fig. 1: supervised, unsupervised, semi-supervised, and reinforcement learning. DL consists of a collection of connected computational neurons, organized in multiple layers[5], between inputs and outputs, the whole capable of learning more complex functions than a single neuron[6] or layer. Each layer extracts some features from its inputs, and each subsequent layer extracts features out of the previous, lower-level features. In this sense, DL extracts high-level latent features (e.g., a human face, a cat, a sentiment) from lower-level features and data (e.g., corners, contours, pixels, letters). The idea of a nested hierarchy of features, each one defined in terms of simpler ones is the essential pillar from which DL derives its great power and flexibility. The *depth* in DL refers to the presence of *hidden layers* in the network between the inputs and the outputs, the results of their computations are not seen by the user or at the output. It was the challenges faced by *shallow* machine learning algorithms[7] and their failure to address central tasks in AI, such as object recognition in computer vision or speech recognition (high-dimensional data), that led to the developments and breakthroughs in DL. To date, DL technology is ubiquitous in smart phones for example, in natural language processing, machine translation, image processing, and recommender systems used by Netflix, Amazon, and YouTube for example. More generally, problems with high-dimensional data in which the curse of dimensionality is a major impediment—and there are scores of them in reliability and safety applications, for example in prognostics and health management—are candidates for being effectively handled by DL

---

[5] With different network architectures, feedforward (DNN) or with feedback looks (recurrent neural networks or RNN).
[6] Typically executing the sigmoid function, the tanh, the softmax function, or more recently the rectified linear unit (ReLU) and the leaky ReLU.
[7] Learning algorithms that have no hidden layers (no depth), which in essence means learning directly from the features in the data without a hierarchical composition of new high-level features from lower-level ones.

algorithms:

> "*Machine learning problems become exceedingly difficult when the number of dimensions in the data is high. This phenomenon is known as the curse of dimensionality. Of particular concern is that the number of possible distinct configurations of a set of variables increases exponentially as the number of variables increases. [Making the mild assumption that the data was generated by] a composition of factors, possibly at multiple levels in a hierarchy allows an exponential gain [...]. The exponential advantages conferred by the use of [deep learning] counter the exponential challenges posed by the curse of dimensionality.*" [17]

A high-level overview of DL can be found in Ref. [150]. The applications of DL in reliability and safety are steadily growing, in anomaly detection for example, fault classification, RUL estimation, maintenance planning, and more broadly in PHM. Some of these applications were discussed in the previous subsections and will not be repeated here. A few others are included here for illustrative purposes. For example, Tamilselvan and Wang [90] developed a deep belief network (DBN) to perform system health diagnostic and classify different states of the system. The authors benchmarked the DBN classification performance against other shallow ML methods such as SVM, and they reported superior classification accuracy. For unsupervised learning, using approaches such as SOM and autoencoder, DL is highly capable of handling high-dimensional and time sequential clustering and anomaly detection. For example, Reddy, et. al. [120] developed a multi-modal deep auto-encoder approach to realize unsupervised anomaly detection of flight raw time series data. The authors validated their deep autoencoder on the NASA DASHlink open database and reported excellent anomaly detection accuracy.

As noted previously, DL algorithms are capable of handling high-dimensional input data. Furthermore, when equipped with convolutional layers, CNN is highly effective in handling extreme high-dimensional data sources such as images and videos. Li et. al. [74] developed CNN model to perform RUL prediction. The authors validated their model performance using NASA's turbofan engine degradation dataset [151, 152] with multiple sensor signals as input. They reported higher accuracy and lower root mean square error (RMSE) compared with other shallow ML methods. Some applications of DRL were noted in the previous subsection in connection with inspection and maintenance planning. An excellent survey of DL applications in PHM is provided in Fink et al. [153].

In all applications to date, DL has significantly outperformed shallow ML algorithms. While this need not reflect a universal truth, we believe the return on investment in DL methods for reliability and safety application is far more likely to yield more dividends than any other shallow methods.

## 4. Future opportunities: promising reliability and safety applications of ML

The previous section provided an overview of the large but fragmented literature on ML in reliability and safety applications. This ever-expanding analytical landscape can be overwhelming to navigate and integrate into a coherent whole. We sought to facilitate this task by "looking back" and providing a scaffolding for, and a wide panoramic view of this field of applications and its major themes. In this section, we "look ahead" and outline some future opportunities for ML in reliability and safety applications. We have already touched on this in the previous section, for example when discussing Safe Reinforcement Learning and variational autoencoders. Some themes or domain of applications discussed next are likely to touch on ongoing efforts by researchers, others are more speculative in nature and are included here because of their potential to help advance our common **reliability and safety agenda: to engineer more reliable and safer systems, cost effectively, to advance accident and injury prevention, and to make the world a safer place** [8]. The tone hereafter is conversational, and the discussion high-level. We restrict the discussion to our (subjective) Top 5 list of most promising opportunities and themes for **leveraging ML in service of this reliability and safety agenda**. This is by design a short, non-exhaustive list intended not to further prolong this work and to invite other researchers to reflect on our selection and complement it with theirs:

1. **Uncertainty quantification and reduction, and deployment considerations:** in all the previous discussions of ML applications, whether RUL prediction, anomaly detection, fault classification, or more generally any type of PHM activity, improving the accuracy of the prediction was a main driver for the use of ML. For anomaly detection or classification, reducing false alarm rates (or misclassification) and missed detection is paramount for a broader adoption of ML. These considerations will continue to be central to ML applications. For meaningful impact, the literature on ML for reliability and safety applications would be well served to address, even if briefly, deployment considerations, which will inevitably lead to issues of accuracy and uncertainty quantification for more informed decision-making. Predicting a single value for an asset RUL, for example, is limited; providing confidence intervals around a RUL expectation is more useful; identifying the sources of uncertainty in the calculation and adopting models that both quantify and reduce the variability in the calculations will be more convincing for the ML technology to be adopted and have a meaningful impact in the field.

---

[8] In the products we design and operate, and more broadly at work, at home, during commutes, and while engaging in leisurely activities.

2. **Machine learning for fleet-level PHM,** and more broadly, **ML for system-of-system PHM**: this theme was identified and discussed by Fink et al. [153]. We repeat it here because we agree with its importance and give it a slightly different twist or complementary perspective. Many of the applications we examined in section 3 addressed PHM issues at the component or subsystem level, ball bearings for example, compression stage in a turbine, induction motors, or batteries. The future challenges of PHM and promising applications are likely to be at a higher level of aggregation, at the fleet level, for example at the level of a mega-constellation of satellites, a fleet of trucks, or a wind turbine farm. More generally, we propose that one particularly fruitful direction for ML is not just in support of PHM at the system level, but more broadly ML for system-of-systems (SoS) reliability and safety applications. This will raise several interesting multidisciplinary challenges, including the need to better integrate ML with systems engineering models and human factors, and to address the variability in configurations and operational conditions of different units in the fleet or systems within the SoS. The advantages and limitations of transfer learning will likely be central in this context. More details on fleet level PHM can be found Fink et al. [153].

3. **Integration of ML with accident databases and Safety Management Systems:** We strongly believe there are many opportunities for teasing out more, better insights from accident databases and safety management systems (SMS) by using ML tools than with traditional statistical tools. These databases can be institutional with government entities or professional societies for example, and SMS are pervasive across all hazardous industries (SMS also includes near miss management systems (NMS), and these are goldmines for ML tools to probe). The next generation of (smart) SMS will likely include ML tools for seamlessly integrating safety-related heterogeneous data collected across different time scales and through different means (e.g., real-time online monitoring, inspections, manually coded NMS data) to improve the efficacy of accident precursor identification and safety interventions. We believe a most promising application of ML is in unleashing its power to harvest more value from near miss data and other accident databases for ultimately improving accident and occupational injury prevention. Data linkages with other non-accident databases can further help provide measures of exposures for determining accident and injury rates, comparing across similar contexts or industries, and identifying predictive features and the strength of their associations with the outcomes for better prevention.

4. **ML integration with wearable computing/sensor for predictive safety analytics:** one particularly useful category of applications for advancing the safety agenda noted previously is the integration of ML with wearable computing/sensors for predictive safety analytics. This can take many forms, and it can address a wide range of safety and injury prevention problems. For example: (i) falls among the elderly is a leading cause of death and injuries among the older adults, 65+, and in the U.S. it leads to more than 2.7 million hospitalizations and 27,000 deaths annually [154]. The associated mental health toll and financial burden are significant. ML and wearable computing are likely to be very helpful in advancing our understanding of this major public health issue and mitigating it to some extent. This is a most worthy topic for ML and safety researchers to contribute to and make a dent in addressing it[9]; (ii) in the U.S. workers sustain over 5,000 fatal occupational injuries and 2.8 million nonfatal injuries every year. Personal protective equipment (PPE) is the last line of defense against adverse consequences of an accident. A smart PPE that integrates wearable sensors can be helpful in reducing this burden of occupational injury. For example, workers at a construction site can wear smart PPE, and their collective streaming data handled by a centralized ML system that performs a safety supervisory function and some safety analytics[10]; (iii) body-cameras coupled with ML image and activity recognition can provide support for technicians during maintenance tasks and inspection. This in turn can be integrated with the permit-to-work system at a company and provide significant safety benefits if properly designed, tested, and rolled out. Beyond these specific examples in (i), (ii), and (iii), we strongly believe there are many fruitful opportunities for exploring applications that integrate ML with wearable computing for predictive safety analytics and injury prevention.

---

[9] In collaboration with epidemiologists and public health researchers.

5. **Deep Gaussian Process (DGP) and Generative Adversarial Network (GAN) for reliability and safety applications:** the previous sections made it clear that there are scores of machine learning models and algorithms, and this wide availability makes it even more important to benchmark the performance of different approaches for model selection. We recommended in 3.1.2 that authors, reviewers, and editors be more mindful of this expectation of transparency for comparative model analysis and selection when writing or reviewing manuscripts in this area of applications. To help researchers focus on more promising ML models for reliability and safety applications, we briefly note in this paragraph two powerful ML models that are likely to (significantly) outperform all others. They make excellent candidates for addressing the uncertainty quantification and reduction issue raised in (1) and all the other applications we listed subsequently. They are the Deep Gaussian process (DGP) and the generative adversarial network (GAN). DGP is a multilayer hierarchical generalization of Gaussian process, and it is formally equivalent to neural networks with infinitely wide hidden layers. The powerful Gaussian Process Regression model is a special case of DGP with a single layer [155]. DGP is a nonparametric probabilistic model with excellent accuracy and better uncertainty estimation and propagation than alternative ML models. Details about this state-of-the-art ML method can be found in Damianou and Lawrence [155]. GAN is another powerful and particularly creative ML generative method developed by Goodfellow et al. [156]. The GAN architecture consists of two neural networks, a generator G and a discriminator D, competing with each other in a classification task for example, the latter (D) trying to accurately classify an object, the former (G) generating (increasingly more) fake objects to trick the discriminator and degrade its performance. It is this antagonism between the discriminator and the generator, or their cat-and-mouse game that confers superior performance to GAN models. The training of GAN, shown in Fig. 5, involves both finding the optimal parameters for the discriminator to maximize its classification of accuracy and those of a generator to maximize the confusion of the discriminator. The training process is iterative until a Nash equilibrium is reached [157].

---

[10] This monitoring might raise privacy concerns, and in addition to the safety focus, industries may also (ab)use it to monitor other aspects of a worker's performance.

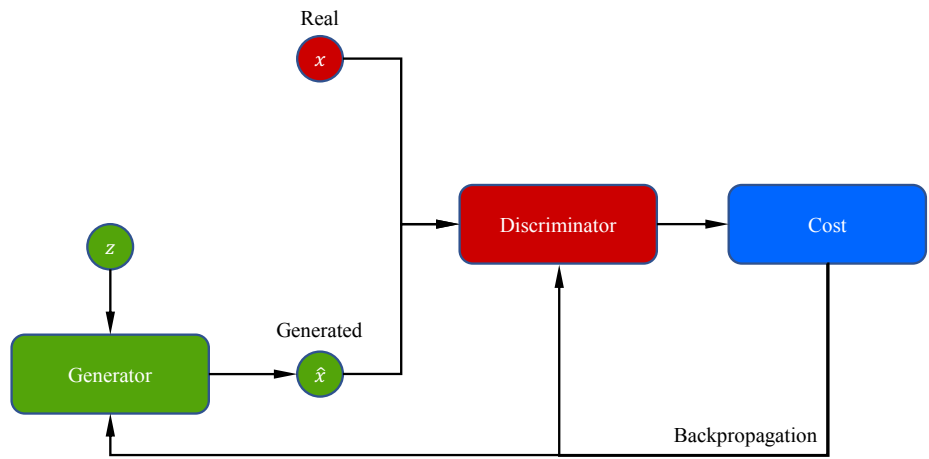

**Figure 5**. Schematic of the GAN training process

Taking the advantages of their excellent accuracy and efficiency in handling high-dimensional data, we anticipate that many promising opportunities will adapt DGP and GAN models for fault detection and classification in a wide range of reliability and safety applications, and more generally that future advanced PHM approaches will leverage these models for more accurate diagnostic, prediction, and uncertainty analysis in complex engineering systems.

We thus conclude our Top 5 list of most promising opportunities and themes for leveraging ML in service of this reliability and safety agenda. There are many other themes and applications we did not include for the sake of brevity, for example the challenges of "small (failure) data" in reliability and safety applications (beyond the familiar "big data" context of ML), the inescapable use of ML for safe autonomous transportation (ground and air), and the need to better integrate ML with risk analysis tools and concepts.

## 5. Conclusion

Machine learning pervades an increasing number of fields. Its impact is profound, and it will continue to upend traditional academic disciplines and industries. Several academic fields have been fundamentally altered by ML, controls and autonomy or computer vision for example, reliability engineering and safety will undoubtedly follow suit. There is already a large but fragmented literature on ML for reliability and safety applications, and it can be overwhelming to navigate and integrate into a coherent whole. In this work, we sought to facilitate this task providing a scaffolding for, and a wide panoramic view of this ever-expanding analytical landscape and highlighting its major landmarks and pathways.

We first provided an overview of the different ML categories and sub-categories or tasks. We then "looked back" and reviewed the use of ML in reliability engineering and safety applications. We selected and discussed a few publications in each category/sub-category, and we devoted a short subsection to the use of Deep Learning in this area to highlight its growing popularity and emphasize its distinctive advantages over other types of ML models. Finally, we "looked ahead" and outlined some promising future opportunities for leveraging ML in service of advancing reliability and safety considerations.

One topic not addressed here (except for the brief discussion of safe reinforcement learning) is illustrated in Fig. 6, and it concerns the reverse relationship of the scope of this work, namely the reliability and safety of ML systems, and more broadly of Artificial Intelligence (AI) systems. This is an important topic, and it deserves careful attention and dedicated treatment in a separate publication. Accidents in ML are defined as "unintended and harmful behavior that may emerge from poor design" of the system [158]. There is a need for new risk analysis methodologies for ML/AI systems, for ways of reducing their accident risks and addressing their fundamental failure mechanisms. An introduction to this topic can be found in Amodei et al. [158], and we propose to examine it more carefully in future work.

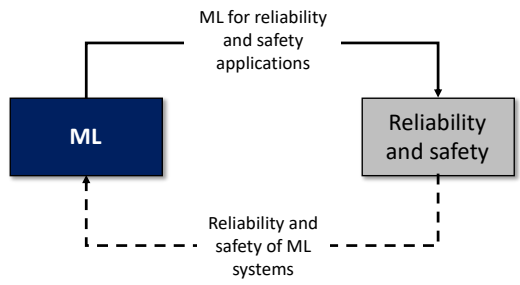

**Figure 6**. ML for reliability and safety application. Not addressed in this work is the reverse of this relationship, namely the reliability and safety of ML systems (and more broadly of AI systems).